\documentclass[letterpaper]{article} 
\usepackage{aaai2026}  
\usepackage{times}  
\usepackage{helvet}  
\usepackage{courier}  
\usepackage[hyphens]{url}  
\usepackage{graphicx} 
\usepackage{natbib}  
\usepackage{caption} 

\usepackage{algorithm}
\usepackage{algorithmic}
\usepackage{amsmath}
\usepackage{booktabs}
\usepackage{newfloat}
\usepackage{listings}
\DeclareCaptionStyle{ruled}{labelfont=normalfont,labelsep=colon,strut=off} 
\floatstyle{ruled}
\newfloat{listing}{tb}{lst}{}
\floatname{listing}{Listing}
\usepackage{amssymb}
\usepackage{pifont}
\newcommand{\cmark}{\checkmark}
\newcommand{\xmark}{\ding{55}}  
\nocopyright
\title{VLHSA: Vision-Language Hierarchical Semantic Alignment for Jigsaw Puzzle Solving with Eroded Gaps}
\author{
  Zhuoning Xu\textsuperscript{\rm 1},
  Xinyan Liu\textsuperscript{\rm 1}
}

\affiliations{
  \textsuperscript{\rm 1}Department of Electrical and Electronic Engineering, The Hong Kong Polytechnic University\\
  11 Yuk Choi Rd, Hung Hom, Kowloon-999077, Hong Kong SAR\\
  rowena.liu@connect.polyu.hk, 
    zx2094@nyu.edu
}

\usepackage{bibentry}

\begin{document}

\maketitle

\begin{abstract}

Jigsaw puzzle solving remains challenging in computer vision, requiring an understanding of both local fragment details and global spatial relationships. While most traditional approaches only focus on visual cues like edge matching and visual coherence, few methods explore natural language descriptions for semantic guidance in challenging scenarios, especially for eroded gap puzzles. We propose a vision-language framework that leverages textual context to enhance puzzle assembly performance. Our approach centers on the Vision-Language Hierarchical Semantic Alignment (VLHSA) module, which aligns visual patches with textual descriptions through multi-level semantic matching from local tokens to global context. Also, a multimodal architecture that combines dual visual encoders with language features for cross-modal reasoning is integrated into this module. Experiments demonstrate that our method significantly outperforms state-of-the-art models across various datasets, achieving substantial improvements, including a 14.2 percentage point gain in piece accuracy. Ablation studies confirm the critical role of the VLHSA module in driving improvements over vision-only approaches. Our work establishes a new paradigm for jigsaw puzzle solving by incorporating multimodal semantic insights.

\end{abstract}


\section{Introduction}

Jigsaw puzzle solving is a classic challenge in computer vision that extends far beyond recreational applications. From archaeological artifact reconstruction~\cite{DERECH2021108065,gigilashvili2023computational} to medical image restoration~\cite{ni2024drcl}, the ability to reassemble fragmented visual content has profound implications across diverse domains. Despite decades of research progress~\cite{Bridger_2020_CVPR}, the accurate reconstruction of jigsaw puzzles remains a difficult task, particularly when dealing with eroded gaps where traditional edge-matching approaches fail~\cite{10.1007/s10044-025-01484-z}.

Most existing methods focus on puzzles without gaps, relying on local visual cues such as edge compatibility~\cite{10.1007/s10044-025-01484-z}, color consistency~\cite{greedy2015,chung1998jigsaw}, and texture patterns to guide assembly. However, the presence of gaps disrupts these local features, making it necessary to incorporate global structural information for accurate reconstruction~\cite{paumard2018imagereassemblycombiningdeep}. 

Recent advances in deep learning have brought significant improvements to puzzle solving through sophisticated neural architectures. Methods like Deepzzle~\cite{Paumard_2020} leverage Siamese networks for pairwise relation learning, while approaches such as PDN-GA~\cite{10096300} employ discriminative networks to validate global arrangements. More recent work has explored transformer-based architectures like ERL-MPP~\cite{Song_Yang_Yao_Ren_Bai_Chen_Jiang_2025} and generative models such as JPDVT~\cite{Liu_2024_CVPR}, treating puzzle assembly as a conditional generation task. However, these vision-only approaches remain limited when visual information alone is insufficient to determine the correct placements.

A key observation motivating the development of the VLHSA model is that, in the process of solving jigsaw puzzles, humans seldom rely solely on visual pattern matching—particularly when confronted with damaged fragments or inter-piece gaps. People usually have a rough idea of what the entire picture is supposed to look like, and this helps them determine where each piece should fit. For instance, when reconstructing a fragmented artwork, knowledge such as “A woman wearing a crown seated on a chair” or “A painting depicts a landscape with mountains and trees” provides strong constraints that effectively supplement visual edge matching.

The emergence of large-scale vision-language models presents a unique opportunity to incorporate such semantic guidance into puzzle assembly~\cite{lyu2025jigsaw}. Recent advances in multimodal understanding have demonstrated remarkable capability in bridging visual content and natural language descriptions, opening new avenues for semantically-informed puzzle reconstruction~\cite{li2025perception, zhu2024vision, han2022survey}. However, effectively integrating language guidance into gap-based puzzle solving requires careful consideration of multiple technical challenges~\cite {kesseli2025logic}.
\begin{figure*}[t]
    \centering
    \includegraphics[width=0.95\textwidth]{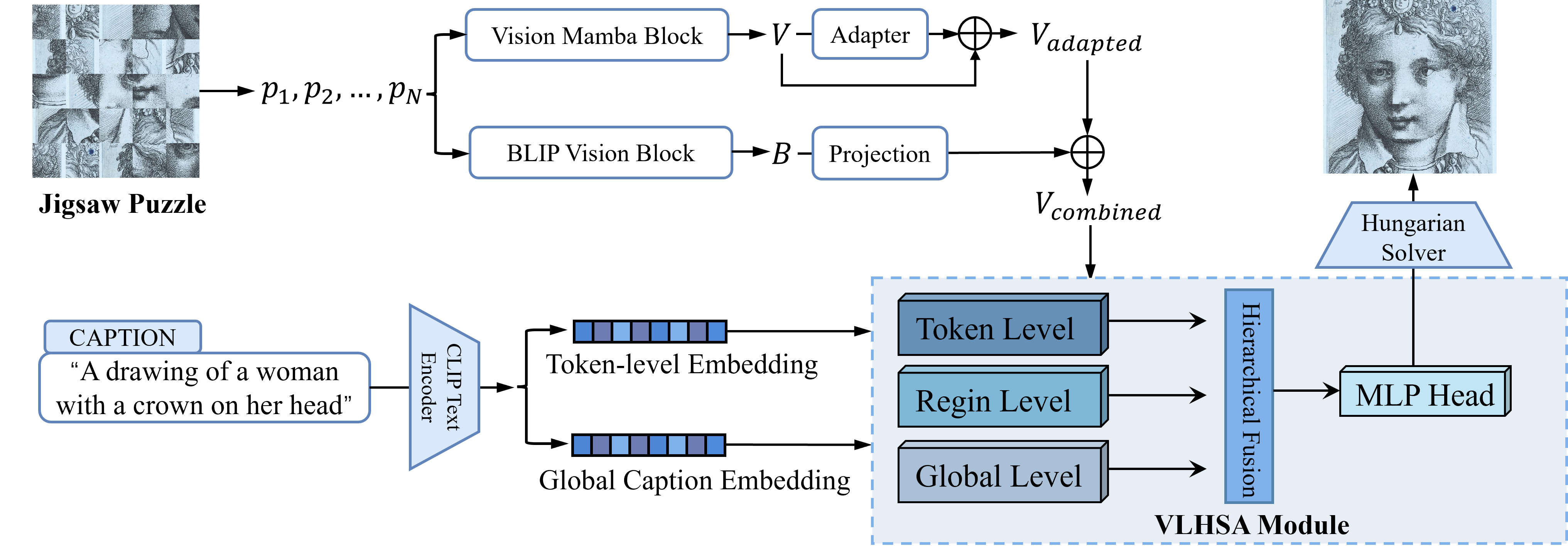} 
    \caption{Overall framework structure of VLHSA for solving jigsaw puzzles with fragment gaps. The model integrates visual features from Vision Mamba and BLIP, while CLIP captures semantic information from captions to guide the alignment. Hierarchical correspondence between visual and textual features is established at the token, region, and global levels. These fused multimodal representations are used to predict piece positions, with the optimal assignment determined by the Hungarian algorithm.}
    \label{fig:framework_overview}
\end{figure*}
To realize this potential, we strategically combine three complementary technologies: Vision Mamba captures long-range spatial dependencies across puzzle pieces through its selective state space mechanism~\cite{zhu2024vision,gu2023mamba}; BLIP provides both semantic caption generation (approximately 10 words each, with manual editing to address model-generated errors and improve clarity) and cross-modally aligned visual features~\cite{li2022blipbootstrappinglanguageimagepretraining,li2023blip}; and CLIP ensures robust text encoding at both global and token levels~\cite{clip}. This combination allows our framework to handle different levels of meaning without slowing down the computation.

Despite these advances, realizing the full benefits of language guidance in gap-based puzzle solving still demands overcoming several open challenges~\cite{lyu2025jigsaw}.To be effective, the model needs to connect small parts of the image with specific words, as well as understand the overall scene described by the text~\cite{wang2025enhancingvisuallanguagemodalityalignment, zhou2024aligningmodalitiesvisionlarge}. The integration of multimodal information must be carefully balanced to complement rather than overshadow visual cues.

Our VLHSA framework addresses these challenges through a hierarchical semantic alignment strategy that establishes correspondences at the token, region, and global levels. This hierarchical strategy allows the model to capture both fine details and the bigger picture, making it easier to place pieces correctly when some visual information is lost or ambiguous.

Figure~\ref{fig:framework_overview} illustrates the overall pipeline of our approach. Given a scrambled puzzle and its semantic caption, we first extract visual features using Vision Mamba and BLIP encoders in parallel, while encoding the caption at both global and token levels with CLIP. The core VLHSA module then performs hierarchical alignment across three semantic granularities, fusing the aligned multimodal features. Finally, the Hungarian algorithm generates the optimal piece-to-position assignment, producing the reconstructed puzzle.

\paragraph{Our contributions can be summarized as follows:}
\begin{itemize}
    \item We are among the first to introduce semantic features into jigsaw puzzle assembly models, demonstrating that textual semantic guidance can significantly enhance performance compared to existing visual-only methods
    
    \item  We propose the VLHSA module, which aligns visual and textual features at token, region, and global levels, facilitating the assembly of puzzles with eroded gaps.

    \item Our method achieves state-of-the-art results on two JPwLEG datasets, while requiring fewer parameters and less training time.
    
\end{itemize}

\section{Related Work}
\paragraph{Optimization-Based Methods}
Optimization-based methods approach puzzle solving as combinatorial optimization problems. Traditional greedy~\citep{greedy2015,Pomeranz} and tabu search~\citep{Adamczewski_2015_ICCV} algorithms offer computational efficiency but typically achieve suboptimal global solutions due to their heuristic nature. More sophisticated approaches, such as nonconvex quadratic programming combined with the projected power method~\citep{FangYan}, and graph-based synchronization methods utilizing the Graph Connection Laplacian~\citep{VahanHuroyan}, leverage global optimization and spectral techniques, thereby overcoming issues with local minima. Nevertheless, all of these methods belong to the category of traditional approaches, lacking consideration of global perception.

\paragraph{Pairwise Relation Learning}
Pairwise relation learning methods model the relationships between adjacent puzzle pieces to facilitate reassembly. \textbf{Deepzzle}~\cite{Paumard_2020} employs a Siamese network~\cite{bertinetto2016fully} to classify relative positions between a central fragment and neighboring pieces, effectively capturing local compatibility but often failing at global coherence, especially for ambiguous or eroded fragments. \textbf{SD$^{2}$RL}~\citep{Song_Jin_Yao_Wang_Ren_Bai_2023} extends this by integrating deep reinforcement learning~\cite{li2017deep}, modeling the puzzle as a Markov Decision Process (MDP). SD$^{2}$RL leverages a Deep Q-Network (DQN)~\cite{mnih2015human}, driven by pairwise adjacency probabilities from Siamese discriminant networks, which significantly improves global spatial coherence through end-to-end learning and sophisticated reward mechanisms. Nonetheless, its scalability to large puzzles remains constrained by computational complexity.

\paragraph{Discriminator-Based Approaches}
Discriminator-based methods utilize neural networks to validate local and global puzzle arrangements. \textbf{PDN-GA}~\citep{10096300} introduces dual discriminative networks combined with genetic algorithms to maximize visual coherence, effectively integrating deep learning with combinational optimization. Although robust, it is computationally intensive and struggles with puzzles containing significant gaps. \textbf{SJP-ED}~\citep{Bridger_2020_CVPR} employs GAN-based inpainting to address boundary erosion issues, but its effectiveness declines when strong global semantic reasoning is required.

\paragraph{Multi-Scale Structural Modeling}
Multi-scale structural methods aim to capture both local and global puzzle structures simultaneously. \textbf{ERL-MPP}~\citep{Song_Yang_Yao_Ren_Bai_Chen_Jiang_2025} utilizes Multi-head Puzzle Perception Networks (MPPN), combining Swin Transformers~\cite{liu2021swintransformerhierarchicalvision} for multi-scale adjacency modeling and evolutionary reinforcement learning for enhanced global optimization. Despite its excellent performance on challenging datasets, its heavy reliance on visual cues makes it susceptible to ambiguity when handling visually similar fragments.

\paragraph{Generative Models}
Generative models have redefined puzzle solving as a conditional image generation task. \textbf{JPDVT}\cite{Liu_2024_CVPR} employs conditional denoising diffusion probabilistic models~\cite{ho2020denoising} combined with Vision Transformers (ViT)~\cite{dosovitskiy2020image}, enabling effective handling of puzzles with missing or large numbers of pieces through generative reconstruction. Similarly, GANzzle++\citep{TALON202535} leverages generative adversarial networks (GAN), Slot Attention, and Vision Transformers for local-to-global assignment in latent spaces, improving robustness, but encountering challenges in computational efficiency. Earlier self-supervised approaches~\citep{doersch2016unsupervisedvisualrepresentationlearning, noroozi2017unsupervisedlearningvisualrepresentations} use context prediction tasks, but their practical applicability diminishes with significant gaps and erosion.

\paragraph{Vision-Language Integration}
Vision-Language Integration has recently advanced in vision-language models like CLIP~\citep{clip} and BLIP~\citep{li2022blipbootstrappinglanguageimagepretraining}, which have achieved remarkable success in cross-modal tasks such as image captioning and visual question answering. While some recent works~\cite{johnson2025hierarchicalvisionlanguagealignmenttexttoimage,HuimingXie,NEURIPS2022_e9882f7f} have begun to explore semantic alignment for vision-language understanding, the use of such semantic guidance for gap-based jigsaw puzzle solving remains largely unexplored. Our VLHSA framework addresses this limitation by systematically leveraging hierarchical vision-language alignment to guide puzzle assembly in challenging scenarios characterized by fragment erosion and gaps.

\section{Methodology}

\subsection{Problem Formulation}

The jigsaw puzzle-solving task with fragment gaps presents a fundamental challenge in computer vision, where the objective is to reconstruct an original image from a collection of disordered pieces. Let $\mathcal{I} = \{p_1, p_2, \ldots, p_N\}$ represent a set of $N$ puzzle pieces extracted from an original image $I_{\text{orig}}$, where each piece $p_i$ corresponds to a distinct spatial region. 

The puzzle reconstruction problem can be formulated as learning an optimal assignment function $f: \mathcal{I} \rightarrow \mathcal{G}$, where $\mathcal{G}$ represents the target grid positions. The objective is to discover the optimal assignment $\sigma^* \in \Sigma$ that minimizes the overall loss:
\begin{equation}
\sigma^* = \arg\min_{\sigma}\ \mathcal{L}r(\sigma)\ +\ \lambda_1 \mathcal{L}s(\sigma)\ +\ \lambda_2 \mathcal{L}_{\text{p}}(\sigma)\ ,
\end{equation}
where $\sigma$ denotes the assignment function mapping puzzle pieces to grid positions; $\mathcal{L}r(\sigma)$ measures spatial reconstruction quality; $\mathcal{L}s(\sigma)$ enforces semantic consistency with the textual description $\mathcal{T}$; $\mathcal{L}_{\text{p}}(\sigma)$ is a pairwise adjacency regularization term; and $\lambda_1$, $\lambda_2$ are balancing weights.

The key insight underlying our approach is that natural language descriptions provide semantic constraints that can guide puzzle assembly beyond purely visual cues. Given a textual description $\mathcal{T}$ that captures the semantic essence of the original image, our model aims to establish meaningful correspondences between visual fragments and linguistic concepts through hierarchical alignment.

\subsection{Framework Architecture}

Our VLHSA framework comprises five synergistic components that collectively address the multimodal puzzle-solving challenge. The Dual Visual Feature Extraction module combines Vision Mamba and BLIP encoders for a comprehensive visual representation. The Multimodal Feature Projection module aligns different feature spaces. The Vision-Language Hierarchical Semantic Alignment (VLHSA) module forms the core innovation, establishing correspondences across multiple semantic granularities. The Feature Integration module combines aligned representations through adaptive fusion mechanisms. Finally, the Assignment Prediction module generates the optimal piece-to-position mapping.

The framework operates through a carefully orchestrated pipeline: visual features are extracted using both Vision Mamba and BLIP vision encoder in parallel, then projected to a unified embedding space. The VLHSA module performs hierarchical alignment across token, region, and global levels using the combined visual features. Aligned features undergo adaptive fusion with learnable weights before being processed by the prediction head to generate the final assignment.

\subsection{Multimodal Feature Extraction}

A dual-encoder architecture that leverages the complementary strengths of Vision Mamba and BLIP vision encoder is employed. This design captures both sequential spatial dependencies and cross-modally aligned visual semantics.

\subsubsection{Vision Mamba Encoding}
Given an input puzzle configuration, we partition it into non-overlapping patches $\{x_1, x_2, \ldots, x_N\}$ and process them through Vision Mamba blocks to obtain sequential visual features $V \in {R}^{N \times d_v}$. To enhance puzzle-specific pattern recognition, we introduce a residual adapter~\cite{rebuffi2017learning}:
\begin{equation}
    V_{\text{adapted}} = V + \mathrm{adapter}(V),
\end{equation}
where the adapter consists of a multi-layer perceptron designed to capture visual patterns relevant for puzzle solving.

\subsubsection{BLIP Visual Encoding}
In parallel, we extract semantically rich patch features using BLIP's vision encoder, which inherently provides vision-language-aligned representations. For each puzzle piece $p_i$, BLIP features $B_i$ are obtained through the pre-trained BLIP vision model, yielding $B \in {R}^{N \times d_b}$.

\subsubsection{Feature Space Alignment}
To enable effective fusion, BLIP features are projected to match Vision Mamba's dimensionality:
\begin{equation}
    B_{\text{proj}} = B W_b + b_{\text{bias}},
\end{equation}
where $W_b \in {R}^{d_b \times d_v}$ and $b_{\text{bias}} \in {R}^{d_v}$ are learnable parameters.

\subsubsection{Dual Feature Integration}
The final visual representation combines both encoders through residual connection:
\begin{equation}
    V_{\text{combined}} = V_{\text{adapted}} + B_{\text{proj}},
\end{equation}
This design preserves spatial structure from Vision Mamba while incorporating semantic richness from BLIP.

\subsubsection{Textual Semantic Encoding}

For textual guidance, we generate concise descriptions of original images using BLIP's captioning capability and extract complementary textual representations using CLIP's text encoder to match the visual feature space:
\begin{itemize}
    \item \textbf{Global caption embedding} ${C}_{\text{global}} \in {R}^{d_t}$: A single vector representing the entire caption's semantic content through mean pooling.
    \item \textbf{Token-level embeddings} ${C}_{\text{tokens}} \in {R}^{L \times d_t}$: Sequential representations preserving fine-grained semantic information for individual words.
\end{itemize}

\subsection{Vision-Language Hierarchical Semantic Alignment}

The core innovation of our approach lies in the VLHSA module, which establishes correspondences between visual patches and textual descriptions across three hierarchical levels. This multi-granularity strategy enables comprehensive semantic understanding from fine-grained word-patch associations to holistic image-caption alignment.

\subsubsection{Token-Level Alignment}

At the finest granularity, we align individual visual patches with specific caption tokens, enabling patches to attend to relevant semantic concepts selectively. This is implemented through multi-head cross-attention~\cite{vaswani2017attention} where visual patches serve as queries and caption tokens as keys and values:
\begin{equation}
{V}_{\text{token}} = \text{MultiHead}({V}_{\text{combined}}, {C}_{\text{tokens}}, {C}_{\text{tokens}}),
\end{equation}

To encourage focused and diverse attention patterns, our models introduce a token-level alignment loss:
\begin{equation}
\mathcal{L}_{\text{token}} = -\alpha \times \frac{1}{N} \sum_{i=1}^N \max_j(A_{i,j}) \log\left(\max_j(A_{i,j}) + \varepsilon\right),
\end{equation}
where $N$ is the number of patches,$A_{i,j}$ denotes the attention weight between the $i$-th patch and the $j$-th token, and $\max_j(A_{i,j})$ represents the highest attention weight assigned by the $i$-th patch across all tokens. $\varepsilon$ is a small constant added for numerical stability, and $\alpha$ is a balancing coefficient. This entropy-based loss encourages each patch to form confident alignments with relevant tokens, while also preventing all patches from collapsing their attention onto a single dominant token.
\subsubsection{Region-Level Alignment}

To capture spatial relationships and contextual dependencies, groups of adjacent patches are aligned with multi-word phrases in the caption. This intermediate level bridges fine-grained token matching and global semantic understanding.

We construct region features by grouping patches into overlapping spatial windows:
\begin{equation}
{R}_k = \frac{1}{|W_k|} \sum_{i \in W_k} {V}_{\text{combined},i}
\end{equation}
where $W_k$ represents the $k$-th spatial window (e.g., $2 \times 2$ or $3 \times 3$ neighborhoods) and $V_{\text{combined},i}$ is the feature of the $i$-th patch. Similarly, phrase features are constructed using sliding windows over caption tokens:
\begin{equation}
{P}_j = \frac{1}{|S_j|} \sum_{t \in S_j} {C}_{\text{tokens},t}
\end{equation}
Region-phrase alignment employs cross-attention~\cite{vaswani2017attention} between these constructed features:
\begin{equation}
{V}_{\text{region}} = \text{CrossAttention}({R}, {P}),
\end{equation}
The region-level alignment loss encourages high-quality correspondences:
\begin{equation}
\mathcal{L}_{\text{region}} = -\frac{1}{|\mathcal{R}||\mathcal{P}|} \sum_{r \in \mathcal{R}, p \in \mathcal{P}} \cos({r}, {p}),
\end{equation}
where $|R|$ denotes the number of regions, $|P|$ denotes the number of phrases, and $\cos(r, p)$ represents the cosine similarity between region $r \in R$ and phrase $p \in P$. The negative sign indicates that the loss aims to maximize the similarity between regions and phrases.
\subsubsection{Global-Level Alignment}

At the highest abstraction level, we align overall image semantics with complete caption meaning to ensure semantic coherence. We encode global visual semantics using a transformer encoder:
\begin{equation}
{V}_{\text{global}} = \text{Encoder}({V}_{\text{combined}}),
\end{equation}
Global alignment employs a gating mechanism to fuse visual and textual information adaptively.
\begin{align}
{g} &= \sigma({W}_g[{V}_{\text{global}}; {c}_{\text{global}}^{\text{expand}}] + {b}_g) \\
{V}_{\text{global}}^{\text{aligned}} &= {g} \odot {V}_{\text{global}} + (1-{g}) \odot {c}_{\text{global}}^{\text{expand}},
\end{align}
where $g$ is the gating weight computed by the sigmoid function $\sigma$, $W_g$ is the weight matrix of the gating network, $[V_{\text{global}}; c_{\text{global}}^{\text{expand}}]$ denotes the concatenation of the global visual feature $V_{\text{global}}$ and the expanded global caption feature $c_{\text{global}}^{\text{expand}}$, and $b_g$ is the bias term of the gating network. The operator $\odot$ represents element-wise multiplication (Hadamard product). $c_{\text{global}}^{\text{expand}}$ is the global textual feature expanded to match the dimension of the visual feature.

\subsubsection{Hierarchical Fusion}

We combine aligned features from all three levels using learnable fusion weights:
\begin{equation}
{V}_{\text{fused}} = \sum_{k=1}^{3} \alpha_k {V}_k,
\end{equation}
where $\boldsymbol{\alpha} = \text{softmax}(\boldsymbol{\theta})$ with learnable parameters $\boldsymbol{\theta} \in {R}^3$, ensuring automatic balancing of different alignment contributions.

\subsection{Assignment Prediction and Optimization}

The fused multimodal features are processed through a specialized prediction head implemented as a multi-layer perceptron:
\begin{equation}
{O} = \text{MLP}({V}_{\text{fused}}),
\end{equation}
where ${O} \in {R}^{N \times N}$ represents position logits for each puzzle piece.

Since puzzle solving is inherently a permutation problem requiring bijective piece-to-position mapping, we employ the Hungarian algorithm~\cite{kuhn1955hungarian} during training to find optimal assignments:
\begin{equation}
\sigma^* = \text{Hungarian}(-{O}),
\end{equation}
Our total training objective combines reconstruction accuracy with semantic alignment, and includes minor auxiliary terms for pairwise adjacency and local consistency regularization:
\begin{equation}
\mathcal{L}_{\text{total}}=\mathcal{L}_{\text{assign}} + \lambda(\mathcal{L}_{\text{token}}+\mathcal{L}_{\text{region}} + \mathcal{L}_{\text{global}})+\lambda_{\text{p}}\mathcal{L}_{\text{p}}
\end{equation}
where $\mathcal{L}_{\text{assign}}$ is the cross-entropy loss for the optimal assignment, and $\lambda$ balances reconstruction accuracy with semantic coherence.

For regularization, we introduce a pairwise adjacency loss ($\mathcal{L}_{\text{p}}$), which serves as minor auxiliary terms without affecting the main vision-language alignment framework.

\begin{algorithm}[t]
\caption{Hierarchical Multimodal Alignment for Jigsaw Puzzle Assignment}
\begin{algorithmic}[1]
\REQUIRE Puzzle image $I$, caption text $T$
\ENSURE Optimal assignment $\sigma^*$
\STATE \textbf{Patch Extraction and Visual Encoding:}
\STATE \hspace{1em} Extract patches $\{x_1, x_2, \ldots, x_n\}$ from $I$
\STATE \hspace{1em} $V \leftarrow \text{VisionMamba}(\{x_1, x_2, \ldots, x_n\})$
\STATE \hspace{1em} $V_{\text{adapted}} \leftarrow V + \text{Adapter}(V)$
\STATE \textbf{BLIP Feature Projection:}
\STATE \hspace{1em} $B \leftarrow \text{BLIP}(\text{patches}(I))$
\STATE \hspace{1em} $B_{\text{proj}} \leftarrow W_b B + b_{\text{bias}}$
\STATE \hspace{1em} $V_{\text{combined}} \leftarrow V_{\text{adapted}} + B_{\text{proj}}$
\STATE \textbf{Caption Encoding (CLIP):}
\STATE \hspace{1em} $C_{\text{global}},\, C_{\text{tokens}} \leftarrow \text{CLIP}(T)$
\STATE \textbf{Token-Level Alignment:}
\STATE \hspace{1em} $V_{\text{token}} \leftarrow \text{MultiHeadAttention}(V_{\text{combined}},\, C_{\text{tokens}},\, C_{\text{tokens}})$

\STATE \hspace{1em} $\mathcal{L}_{\text{token}} \leftarrow -\frac{1}{N} \sum_{i} \max_j(A_{i,j}) \log(\max_j(A_{i,j}) + \varepsilon)$
\STATE \textbf{Region-Level Alignment:}
\STATE \hspace{1em} $R \leftarrow \text{ConstructSpatialRegions}(V_{\text{combined}})$
\STATE \hspace{1em} $P \leftarrow \text{ConstructPhrases}(C_{\text{tokens}})$
\STATE \hspace{1em} $V_{\text{region}} \leftarrow \text{CrossAttention}(R, P)$
\STATE \hspace{1em} $\mathcal{L}_{\text{region}} \leftarrow -\frac{1}{|R||P|} \sum_{r,p} \cos(r, p)$
\STATE \textbf{Global-Level Alignment \& Hierarchical Fusion:}
\STATE \hspace{1em} ${V}_{\text{global}}^{\text{aligned}} = {g} \odot {V}_{\text{global}} + (1-{g}) \odot {c}_{\text{global}}^{\text{expand}}$
\STATE \hspace{1em} $V_{\text{fused}} \leftarrow \sum_{i} \alpha_i V_i,\,\,\, \alpha = \text{softmax}(\theta)$
\STATE \hspace{1em} $O \leftarrow \text{MLP}(V_{\text{fused}})$
\STATE \hspace{1em} $\sigma^* \leftarrow \text{Hungarian}(-O)$
\STATE \textbf{Total Loss:}
\STATE \hspace{1em} $\mathcal{L}_{\text{total}} \leftarrow \mathcal{L}_{\text{assign}} + \lambda(\mathcal{L}_{\text{token}}+\mathcal{L}_{\text{region}} + \mathcal{L}_{\text{global}})+\lambda_{\text{p}}\mathcal{L}_{\text{p}}$
\RETURN $\sigma^*$
\end{algorithmic}
\end{algorithm}

\section{Experiments}
\subsection{Datasets}
Our VLHSA framework is evaluated on two challenging jigsaw puzzle datasets with fragment gaps, specifically designed for real-world archaeological reconstruction scenarios.

Datasets are selected based on the following criteria:\
(1) Rich semantic content for effective vision-language alignment;\
(2) Recognized benchmarks in the literature for fair comparison;\
(3) Availability of multiple puzzle sizes ($3\times3$, $5\times5$) to assess scalability.

\subsubsection{JPwLEG-3.}
The JPwLEG-3 dataset~\cite{Song_Jin_Yao_Wang_Ren_Bai_2023} comprises 12,000 images from the MET Museum collection, each segmented into $3\times3$ fragments (96$\times$96 pixels) with 48-pixel gaps. Fragments are randomly displaced by $\pm$7 pixels to simulate misalignment. The standard split is 9,000 for training, 1,000 for validation, and 2,000 for testing.

\subsubsection{JPwLEG-5.}
The JPwLEG-5 dataset~\cite{Song_Jin_Yao_Wang_Ren_Bai_2023} contains 12,000 puzzles, each divided into $5\times5$ pieces with 12-pixel gaps, representing a higher-complexity scenario. The split follows the same protocol: 9,000 for training, 1,000 for validation, and 2,000 for testing.

Both datasets are derived from the MET open-access collection, providing diverse and semantically rich artwork that benefits vision-language based puzzle reconstruction.
\subsection{Experimental Setup}
We strictly follows the data preprocessing and evaluation protocols established in SD$^{2}$RL~\cite{Song_Jin_Yao_Wang_Ren_Bai_2023}. All baseline methods included in our comparison are based on the original implementations and evaluation criteria, as the JPwLEG dataset was first introduced in that work. Our experiments are conducted on a single NVIDIA GeForce RTX 3090 GPU with CUDA version 12.4. The Vision Mamba backbone is configured with a patch size of 96, an embedding dimension of 256, and a depth of 24. The batch size is set to 32 for all training, validation, and testing phases. The optimizer is AdamW~\cite{loshchilov2019decoupledweightdecayregularization} with an initial learning rate of 0.001 and weight decay of $1\times10^{-4}$. We use a cosine annealing learning rate scheduler ($T_{\mathrm{max}}=50$, $\eta_{\mathrm{min}}=1\mathrm{e}{-6}$). The loss function is cross-entropy with label smoothing of 0.08.
All experiments are conducted five times, and we report the average performance.

Performance is measured using four criteria: \textbf{Perfect}, \textbf{Piece}, \textbf{Horizontal}, and \textbf{Vertical} accuracy. Perfect denotes the fraction of puzzles solved without any errors, Piece captures the rate at which pieces are assigned to their original locations, and Horizontal and Vertical refer to the accuracy of restoring correct pairwise relationships along each axis.

\subsection{Comparison Results on JPLEG-5 Dataset}
\renewcommand{\arraystretch}{1.3} 
\setlength{\tabcolsep}{1.05mm} 
\begin{table}[htbp]
\centering
\begin{tabular}{lccccc}
\hline
Method & Venue & Perf. & Piece & Hori. & Vert. \\
\hline
Deepzzle & TIP-2020  & 0.0  & 21.9  & 10.9  & 10.7  \\
SD$^{2}$RL    &AAAI-2023  & 5.1  & 40.3  & 26.5  & 26.2  \\
PDN-GA   &ICASSP-2023& 6.1  & 44.3  & 30.8  & 30.6  \\
ERL-MPP  & AAAI-2025 & 18.6 & 52.7  & 56.5  & 57.3  \\
\textbf{VLHSA (Ours)}& -  & \textbf{19.0} & \textbf{66.9} & \textbf{58.1} & \textbf{58.4} \\
\hline
\end{tabular}
\caption{Comparison of state-of-the-art models on the JPLEG-5 dataset in terms of Perfect, Piece, Horizontal, and Vertical Accuracy (\%).}

\end{table}
\setlength{\tabcolsep}{6pt} 

Table 1 summarizes the comparison results on the JPLEG-5 dataset. Solving jigsaw puzzles with large gaps remains a highly challenging task, especially since many pieces exhibit nearly identical color backgrounds, making it extremely difficult to achieve perfect reconstruction. As shown in the table, while previous methods such as PDN-GA and SD$^{2}$RL only achieve Perfect accuracies of 6.1\% and 5.1\%, respectively, most traditional baselines are unable to reassemble any puzzles perfectly. The recent ERL-MPP model achieves a Perfect Accuracy of 18.6\%.

Our proposed VLHSA framework achieves a new state-of-the-art Perfect Accuracy of 19.0\%, slightly outperforming ERL-MPP. More notably, VLHSA attains a Piece Accuracy of 66.9\%, which is a significant improvement of 14.2 percentage points over ERL-MPP (52.7\%). In addition, our approach achieves 58.1\% and 58.4\% on Horizontal and Vertical Accuracy, respectively, demonstrating consistent gains in both global and local spatial relationships. 

These results demonstrate the robustness of our method, especially for high piece similarity, and confirm that hierarchical vision-language alignment offers clear improvements over leading vision-based baselines.
\setlength{\tabcolsep}{1.5mm} 
\begin{table}[htbp]
\centering
\begin{tabular}{lccccccc}
\hline
Category & Perfect  & 2-off & 3-off & 4-off & 5-off & $\geq$6-off \\
\hline
Percentage        & 19.00    & 10.90 & 4.00  & 7.10  & 3.85  & 55.15 \\
\hline
\end{tabular}
\caption{Distribution of reconstruction error levels on the JPLEG-5 test set.}
\end{table}

Table 2 illustrates the distribution of reconstruction errors on the JPLEG-5 test set. The model achieves a perfect accuracy of 19.0\%, with 10.9\% of puzzles solved within two misplaced pieces. The majority of errors occur in more challenging cases, particularly those involving fragments with highly similar visual appearances.
\renewcommand{\arraystretch}{1.1}
\setlength{\tabcolsep}{3.05mm}
\begin{table}[htbp]
\centering
\begin{tabular}{lcccc}
\hline
Method        & Pnt. (\%)     & Eng. (\%)    & Art. (\%)  \\
\hline
Deepzzle & 15.5        & 23.9         & 26.3    \\
SD$^{2}$RL    & 23.8        & 48.5         & 48.6     \\ 
PDN-GA    & 24.5        & 54.5         & 54.0     \\
ERL-MPP   & 28.5        & 62.8         & 66.8    \\
\textbf{VLHSA (Ours)}    & \textbf{44.7} & \textbf{77.9} & \textbf{78.2} \\
\hline
\end{tabular}
\caption{Piece Accuracy (\%) across different categories in the JPLEG-5 dataset. VLHSA outperforms all state-of-the-art models in every category.}
\end{table}

Table 3 presents the Piece Accuracy results grouped by puzzle type. The dataset includes three major forms of visual content—artworks resembling traditional paintings (Pnt.), detailed engravings (Eng.), and structured artifact images (Art.). Across all categories, our VLHSA model consistently achieves the highest accuracy, notably outperforming previous state-of-the-art methods. The largest improvements are observed on engravings and artifacts, where VLHSA exceeds the strongest baseline (ERL-MPP) by over 15 percentage points, demonstrating its robustness in handling fine-grained structural and visual cues.
\subsection{Comparison Results on JPLEG-3 Dataset}
\setlength{\tabcolsep}{2.2mm}
\renewcommand{\arraystretch}{1.1}
\begin{table}[htbp]
\centering
\begin{tabular}{lcc}
\hline
Method & Venue & Piece Accuracy (\%) \\
\hline
Deepzzle& TIP-2020 & 52.3 \\
SJP-ED& CVPR-2020 & 55.2 \\
PDN-GA& ICASSP-2023 & 58.2 \\
JPDVT  & CVPR-2024 & 71.3 \\
SD$^{2}$RL& AAAI-2023 & 81.6 \\
\textbf{VLHSA (Ours)}& - & \textbf{85.4} \\
\hline
\end{tabular}
\caption{Comparison of Piece Accuracy (\%) on the JPwLEG-3 dataset. VLHSA achieves the best result among all baselines.}
\end{table}
On the JPwLEG-3 dataset, which consists of $3\times3$ puzzles with lower structural complexity, VLHSA still achieves the highest Piece Accuracy of 85.4\%, outperforming all prior methods including SD$^{2}$RL (81.6\%). This result confirms the model’s strong generalization ability across different puzzle scales.
\begin{table}[htbp]
\centering
\renewcommand{\arraystretch}{1.1}
\setlength{\tabcolsep}{3mm}
\begin{tabular}{lccc}
\hline
Method     & Pnt. (\%) & Eng. (\%) & Art. (\%) \\
\hline
Deepzzle & 30.6          & 58.9           & 67.4          \\
Greedy   & 33.7          & 61.9           & 70.0          \\
Tabu     & 33.1          & 63.9           & 68.6          \\
GA       & 33.9          & 63.1           & 70.4          \\
SD$^{2}$RL & 37.3          & 67.5           & 74.3          \\
\textbf{VLHSA (Ours)} & \textbf{73.5} & \textbf{92.7}  & \textbf{89.9} \\
\hline
\end{tabular}
\caption{Category-wise Piece Accuracy (\%) comparison on the JPwLEG-3 dataset. Our VLHSA method achieves substantial improvements over all previously published baselines. }
\end{table}

Our proposed VLHSA method significantly outperforms all previous baselines across all categories on the JPwLEG-5 dataset. Specifically, VLHSA achieves 73.5\% on Pnt., 92.7\% on Eng., and 89.9\% on Art., representing substantial improvements compared to the best-performing baseline (SD$^{2}$RL), which only achieves 37.3\%, 67.5\%, and 74.3\% respectively.

Across both the 5×5 and 3×3 puzzle settings, VLHSA exhibits clear and consistent advantages over existing approaches. Notably, its performance gains on the more complex 5×5 configuration underscore the method’s robustness in handling increased structural ambiguity and visual similarity. Meanwhile, the strong results on 3×3 puzzles indicate that these improvements are not limited to specific puzzle scales, but rather reflect a generalizable modeling capability. Collectively, these findings demonstrate that hierarchical vision-language alignment offers substantial benefits for jigsaw reassembly, independent of puzzle difficulty or content type.

\subsection{Ablation Study}

\begin{table}[htbp]
\centering
\renewcommand{\arraystretch}{1.1}
\setlength{\tabcolsep}{1.7mm}
\begin{tabular}{ccc|cccc}
\hline
\textbf{Mamba} & \textbf{BLIP} & \textbf{CLIP} & \textbf{Perf.} & \textbf{Piece} & \textbf{Hori.} & \textbf{Vert.} \\
\hline
\cmark & \xmark & \xmark & 3.55  & 49.29 & 39.39 & 39.09 \\
\cmark & \cmark & \xmark & 5.95  & 53.98 & 43.63 & 43.53 \\
\cmark & \cmark & \cmark & 8.85  & 58.57 & 48.64 & 48.50 \\
\hline
\end{tabular}
\caption{Ablation study on encoder components conducted on the JPwLEG-5 dataset. \cmark\ indicates the module is used. Adding BLIP and CLIP progressively improves accuracy.}
\label{tab:encoder_ablation}
\end{table}

\begin{table}[htbp]
\centering
\renewcommand{\arraystretch}{1.1}
\setlength{\tabcolsep}{1.55mm}
\begin{tabular}{ccc|cccc}
\hline
\textbf{Token} & \textbf{Region} & \textbf{Global} & \textbf{Perf.} & \textbf{Piece} & \textbf{Hori.} & \textbf{Vert.} \\
\hline

\xmark & \xmark & \cmark & 17.50 & 66.43 & 57.27 & 57.16 \\
\cmark & \xmark & \cmark & 17.75 & 66.13 & 57.16 & 57.01 \\
\xmark & \cmark & \cmark & 17.30 & 66.00 & 56.50 & 56.58 \\
\cmark & \cmark & \cmark & \textbf{19.00} & \textbf{66.90} & \textbf{58.05} & \textbf{58.43} \\
\hline
\end{tabular}
\caption{Ablation of VLHSA alignment modules under full encoder setup on JPwLEG-5 dataset. \cmark\ indicates enabled module. Global alignment contributes the most, while token and region provide complementary gains.}
\label{tab:vlhsa_module_ablation}
\end{table}

We conduct ablation studies in two stages to assess the contributions of encoder components and semantic alignment modules. As shown in Table~\ref{tab:encoder_ablation}, the Vision Mamba alone yields limited performance, with only 3.55\% Perfect Accuracy. Introducing BLIP improves both global and local placement, and the addition of CLIP further enhances overall accuracy, suggesting that cross-modal features play a critical role in resolving visual ambiguity.

Table~\ref{tab:vlhsa_module_ablation} examines the effect of individual semantic alignment modules. Among them, global alignment contributes the most significantly, boosting Perfect Accuracy to 17.50\% when used alone. Token-level and region-level alignment offer moderate but consistent gains and are more effective when used in conjunction with the global module. The best performance is achieved when all three modules are enabled, reaching 19.00\% Perfect Accuracy and 66.9\% Piece Accuracy. This confirms that hierarchical alignment across different semantic levels is essential for fine-grained spatial reasoning and global structural consistency.

In summary, both the multimodal encoder design and the hierarchical alignment strategy are essential to the effectiveness of our model. The global module provides strong semantic grounding, while token and region modules refine alignment at different spatial scales.

\subsection{Computational Cost Analysis}
\begin{table}[htbp]
\centering
\renewcommand{\arraystretch}{1.1}
\setlength{\tabcolsep}{1.3mm}

\label{tab:cost}
\begin{tabular}{lccc}
\toprule
\textbf{Method} & \textbf{Training(h)} & \textbf{Test(s)} & \textbf{Model Size (M)} \\
\midrule
SD$^2$RL  & 30.7 & 44.7 & 44.2 \\
PDN-GA  & 19.7 & 32.5 & 46.8 \\
ERL-MPP & 36.2 & 53.9 & 102.1 \\
\textbf{VLHSA (Ours)} & \textbf{1.68}$^*$ & \textbf{10.1}$^*$ & \textbf{27.1} \\
\bottomrule
\end{tabular}
\caption{Comparison of computational costs on the JPwLEG-5 dataset. $^*$Evaluated on NVIDIA RTX 3090. Baseline methods do not specify hardware configurations.}
\end{table}
Table 8 compares the computational efficiency of recent state-of-the-art methods. Our VLHSA framework demonstrates superior efficiency with significantly reduced training time, competitive inference speed, and the most compact model size. While baseline methods do not specify their hardware configurations, our results provide a clear reference point for computational requirements.

\section{Conclusion}
This paper introduces a novel vision-language framework for jigsaw puzzle solving with fragment gaps, addressing challenges where traditional visual cues are insufficient. Our key innovation, the Vision-Language Hierarchical Semantic Alignment (VLHSA) module, establishes correspondences between visual fragments and textual descriptions across token, region, and global semantic levels. Experimental results demonstrate significant improvements over state-of-the-art methods on both JPwLEG datasets. Comprehensive ablation studies confirm that hierarchical semantic alignment provides crucial constraints for puzzle assembly, with global alignment contributing most significantly while token and region levels offer complementary fine-grained reasoning. This work establishes a new paradigm for puzzle solving by demonstrating that natural language descriptions can effectively guide spatial reconstruction tasks, opening promising directions for incorporating multimodal reasoning in complex visual assembly problems.

\bigskip

\bibliography{aaai2026}

\newpage
\appendix
\section{Implementation Details}

We conducted all experiments on a single NVIDIA RTX 3090 GPU with CUDA version 12.4 using PyTorch and HuggingFace Transformers. 

The JPwLEG-3 and JPwLEG-5 datasets used in our experiments are directly provided in .npy format by \cite{Song_Jin_Yao_Wang_Ren_Bai_2023} and are publicly available online. All images and permutation labels are pre-generated and fixed in the dataset, rather than being randomly shuffled by us. This approach maintains consistency with prior work and ensures fair comparison across methods. For each sample, we use the preprocessed images and features provided by the dataset authors: JPwLEG-3 contains $288 \times 288$ images divided into $3 \times 3$ patches (each patch is $96 \times 96$), and JPwLEG-5 contains $480 \times 480$ images divided into $5 \times 5$ patches (each patch is $96 \times 96$). Each sample includes the corresponding permutation label (of dimension 9 or 25), BLIP patch-level features, and CLIP-based caption embeddings (both global and token-level).

Image captions were generated automatically and manually verified. Global and token-level textual embeddings were extracted using CLIP (\texttt{vit-base-patch32}), with global features obtained via mean pooling.

During training, we apply color jittering along with ImageNet normalization. Test and validation sets use normalization only.

The vision backbone is Vision Mamba, with a patch size of 96, an embedding dimension of 256, and a depth of 24. BLIP and CLIP features are projected to a unified embedding space. The model includes hierarchical semantic alignment at token, region, and global levels, followed by a multi-layer perceptron for assignment prediction.

The model was trained with AdamW (learning rate $1 \times 10^{-3}$, weight decay $1 \times 10^{-4}$) and label smoothing of 0.02. ReduceLROnPlateau was used for learning rate scheduling. Batch size was 32, with up to 200 training epochs.

Piece-to-position assignment was solved using the Hungarian algorithm. Evaluation metrics include piece accuracy and perfect reconstruction rate. We select the best model checkpoint based on validation performance to avoid test set overfitting.
\section{Visual Comparison of Puzzle Assembly}

\begin{figure}[htbp]
    \centering
    \includegraphics[width=0.48\textwidth]{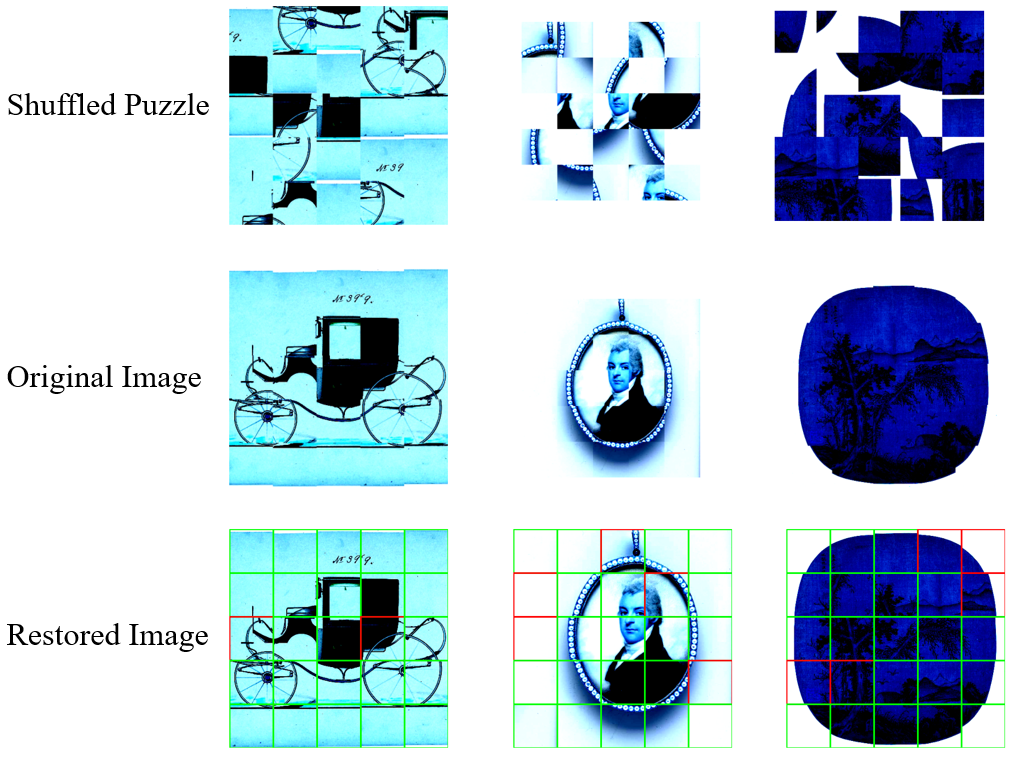}
    \caption{
Qualitative results on JPwLEG-5 dataset. Top row shows scrambled input, middle row shows ground truth, bottom row shows our reconstruction. Green boxes mark correctly placed pieces, red boxes show errors.
    }
    \label{fig:qualitative_jpwleg5}
\end{figure}

Many puzzle pieces in our dataset look remarkably similar, sharing colors, textures, and even small objects. This results in substantial ambiguity for both visual and language-guided models, as many fragments share nearly identical color distributions, textures, or local objects. This makes it nearly impossible to place pieces correctly using only local visual cues. Even with the incorporation of global semantic cues, piece-wise ambiguities remain significant, frequently leading to incorrect assignments in regions with repetitive patterns or backgrounds. The examples showed in Figure 2 below highlight this core challenge - when pieces look so similar, even our multi-modal approach struggles.

Our hierarchical semantic alignment cannot fully resolve cases where pieces look nearly identical. When fragments share similar visual and semantic properties, neither visual nor textual cues provide enough discriminative power.

\begin{figure}[htbp]
    \centering
    \includegraphics[width=0.47\textwidth]{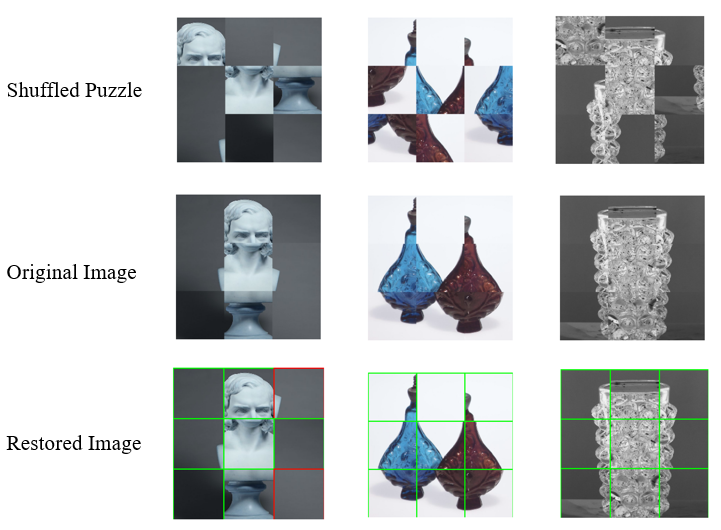}
    \caption{
Qualitative results on JPwLEG-3 dataset. Green boxes mark correctly placed pieces, red boxes show errors.
    }
    \label{fig:qualitative_jpwleg3}
\end{figure}

\begin{table}[!htbp]
\centering
\setlength{\tabcolsep}{3.7mm}
\begin{tabular}{l lcc}

\toprule
Dataset  & Pnt. (\%) & Eng. (\%) & Art. (\%)  \\
\midrule
JPwLEG-3   & 73.5 & 92.7 & 89.9  \\
JPwLEG-5    & 44.7 & 77.9 & 78.2  \\
\bottomrule
\end{tabular}
\caption{
 Piece Accuracy (\%) on JPwLEG-3 and JPwLEG-5. 
}
\label{tab:category_accuracy}
\end{table}
Our VLHSA model performs well on 3×3 puzzles but struggles more with 5×5 puzzles. Engraving images work best in both cases, while paintings drop the most - from 73.5\% to 44.7\% - probably because similar textures and colors make the pieces harder to distinguish.
\end{document}